\documentclass[10pt,twocolumn,letterpaper]{article}

\usepackage{wacv}              

\usepackage{graphicx}
\usepackage[accsupp]{axessibility}
\usepackage{amsmath}
\usepackage{amssymb}
\usepackage{booktabs}

\usepackage{times}
\usepackage{epsfig}
\usepackage[utf8]{inputenc} 
\usepackage[T1]{fontenc}    
\usepackage{algorithm,amsfonts,bm,enumerate,microtype,nicefrac,url,wrapfig,xcolor}
\usepackage{algorithmic}
\usepackage{adjustbox}

\newcommand{\ourModel}{CXDA}

\newcommand{\qx}{\boldsymbol{x}_q}
\newcommand{\sx}{\boldsymbol{x}_s}
\newcommand{\feat}{f_{\boldsymbol{\theta}}}
\newcommand{\featparams}{\boldsymbol{\theta}}
\newcommand{\cls}{f_{\boldsymbol{\phi}}}
\newcommand{\clsparams}{\boldsymbol{\phi}}
\newcommand{\xa}{\texttt{CA}_{\boldsymbol{\omega}}}
\newcommand{\mxa}{\texttt{MCA}_{\boldsymbol{\omega}}}
\newcommand{\xaparams}{\boldsymbol{\omega}}
\newcommand{\keypoint}[1]{\noindent\textbf{#1}\quad}

\usepackage[pagebackref,breaklinks,colorlinks]{hyperref}

\usepackage[capitalize]{cleveref}
\crefname{section}{Sec.}{Secs.}
\Crefname{section}{Section}{Sections}
\Crefname{table}{Table}{Tables}
\crefname{table}{Tab.}{Tabs.}

\def\wacvPaperID{1590} 
\def\confName{WACV}
\def\confYear{2024}

\begin{document}

\title{Feed-Forward Latent Domain Adaptation}

\author{
  Ondrej Bohdal$^1$\thanks{Work done during an internship at Samsung AI Center Cambridge.}, Da Li$^2$, Shell Xu Hu$^2$, Timothy Hospedales$^{1,2}$ \\
  $^1$ School of Informatics, The University of Edinburgh, UK \\
  $^2$ Samsung AI Center Cambridge, UK\\
  \tt\small{\{ondrej.bohdal, t.hospedales\}@ed.ac.uk}, \tt\small{\{da.li1, shell.hu\}@samsung.com} \\
}

\maketitle

\begin{abstract}
We study a new highly-practical problem setting that enables resource-constrained edge devices to adapt a pre-trained model to their local data distributions. Recognizing that device's data are likely to come from multiple latent domains that include a mixture of unlabelled domain-relevant and domain-irrelevant examples, we focus on the comparatively under-studied problem of latent domain adaptation. 
Considering limitations of edge devices, we aim to only use a pre-trained model and adapt it in a feed-forward way, without using back-propagation and without access to the source data. 
Modelling these realistic constraints bring us to the novel and practically important problem setting of feed-forward latent domain adaptation.
Our solution is to meta-learn a network capable of embedding the mixed-relevance target dataset and dynamically adapting inference for target examples using cross-attention. The resulting framework leads to consistent  improvements over strong ERM baselines. We also show that our framework sometimes even improves on the upper bound of domain-supervised adaptation, where only domain-relevant instances are provided for adaptation. This suggests that human annotated domain labels may not always be optimal, and raises the possibility of doing better through automated instance selection.
\end{abstract}

\section{Introduction}
Domain shift presents a real-world challenge for the application of machine learning models because performance degrades when deployment data are not from the training data distribution.
This issue is ubiquitious as it is often impossible or prohibitively costly to pre-collect and annotate training data that are sufficiently representative of test data statistics. The field of domain adaptation \cite{kouw2021daReview,csurka2022visual} has therefore attracted a lot of attention with the promise of adapting models during deployment to perform well using only unlabeled deployment data. The main body of work in deep domain adaptation assumes that there is a pre-specified source domain and a pre-specified target domain. An unlabeled adaptation set is provided from the target domain, and various methods define different learning objectives that update a deep model on the unlabeled adaptation set, with the aim of improving performance on new test data drawn from the target domain. 

In this paper we make two main contributions: A conceptual contribution of a new highly practical variant of the domain adaptation problem; and an algorithm for effective domain adaptation in this condition.

\begin{figure*}[t]
\begin{center}
  \includegraphics[width=0.74\linewidth]{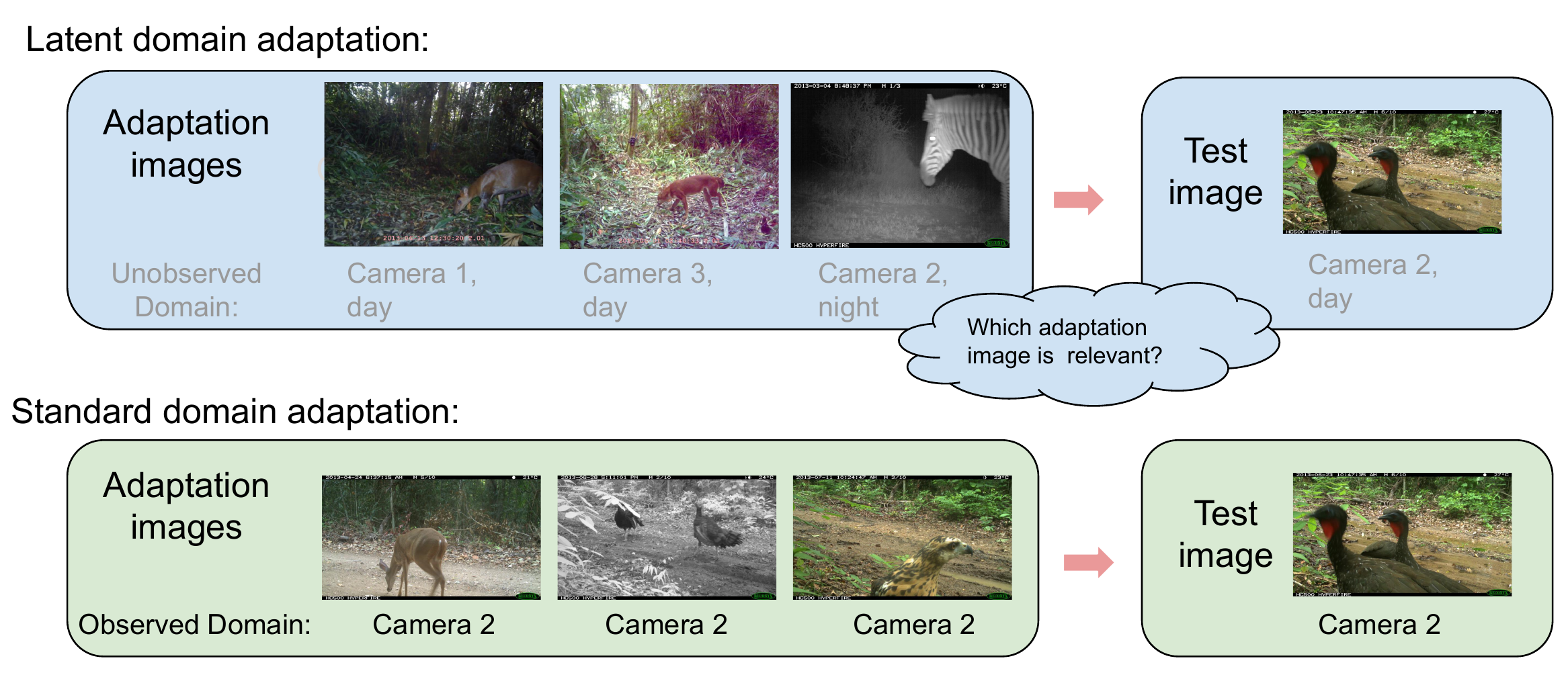}
  \caption{Illustration of standard and latent domain adaptation (LDA) settings. In the LDA setting (support) images come from a variety of domains of mixed and unknown relevance to the test (query) image. In standard DA adaptation images are all assumed to be equally relevant.}
  \label{fig:ldatask}
 \end{center}
\end{figure*}

\keypoint{A motivating scenario} Let us introduce an illustrative application scenario that motivates the variant of the domain adaptation problem that we propose here. Suppose that a robot or other mobile embedded vision system needs to recognise objects. Because it is mobile, it may encounter objects in different unconstrained contexts, e.g. indoor or outdoor backgrounds, sunny or rainy weather, rooms with lights on or lights off. The robot’s object recognition model should adapt to maintain strong performance across all these conditions, for example by adapting based on a buffer of recently experienced unlabelled images. However, unlike standard pre-defined domain adaptation benchmarks with neatly curated domains, there are two new challenges: 1) Using such a buffer as the adaptation set means the adaptation data can be of mixed relevance to the test image to be processed at any given instant. For example, the recent history used for adaptation may span multiple rooms, while any individual test image comes from a specific room. 2) The adaptation needs to happen on-board the robot and ideally happen in real-time as the adaptation set itself is updated over time. The first challenge is the \emph{latent domain} challenge, wherein uncurated adaptation sets do not have consistent relevance to a given test image (Figure~\ref{fig:ldatask}). The second challenge requires adaptation to take place without back-propagation as it is too slow and not supported on most embedded platforms. It means adaptation should be \emph{feed-forward}.

\keypoint{Latent domain adaptation} While domain adaptation is very well studied \cite{kouw2021daReview,csurka2022visual}, most work assumes instances have been pre-grouped into one or more subsets (domains) that differ statistically across groups, while being similar within groups. We join a growing minority \cite{Mancini2021InferringAdaptation,Deecke2022VisualDomains,hoffman2014continuousDA,wang2022continualTTT} in arguing this is an overly restrictive assumption that does not hold in most real applications of interest. Some collection processes may not provide meta-data suitable for defining domain groupings. Alternatively, for other data sources that occur with rich meta-data there may be no obviously correct grouping and existing domain definitions may be sub-optimal \cite{Deecke2022VisualDomains}. Consider the popular iWildCam  \cite{Beery2020TheDataset} benchmark for animal detection within the WILDS \cite{koh2021wilds} suite. The default setup within WILDS defines domains by camera ID. But given that images span different weather conditions and day/night cycles as well as cameras, such domains may neither be internally homogenous, nor similarly distinct. There may be more transferability between images from nearby cameras at similar times of day than between images from the same camera taken on a sunny day vs a snowy night. As remarked by \cite{hoffman2014continuousDA,wang2022continualTTT}, domains may more naturally define a continuum, rather than discrete groups. That continuum may even be multi-dimensional -- such as timestamp of image and spatial proximity of cameras. Our latent domain formulation of the domain adaptation problem spans all these situations where domains are hard to define, while aligning with the requirements of real use cases. 

\keypoint{Feed-forward domain adaptation} Unsupervised domain adaptation aims to adapt models from source datasets (e.g. ImageNet) to the peculiarities of target data distributions in the wild. The mainstream line of work updates models by back-propagation on an adaptation set from the target data distribution \cite{kouw2021daReview,csurka2022visual} (and often simultaneously uses the source data \cite{Liang2020DoAdaptation}). We consider adaptation under the practical constraints of an edge device, namely that neither the hardware capability nor the software stack support back-propagation. Therefore we focus on the \emph{feed-forward} case where adaptation algorithms should use only feed-forward operations, and only the target dataset (source-free condition \cite{Liang2020DoAdaptation}). For example, simply updating batch normalization statistics, which can be done without back-propagation, provides a strong baseline for back-propagation-free adaptation \cite{Schneider2020ImprovingAdaptation,Zhang2021AdaptiveShift}. 

\begin{figure}[t]
  \centering
  \includegraphics[width=\columnwidth]{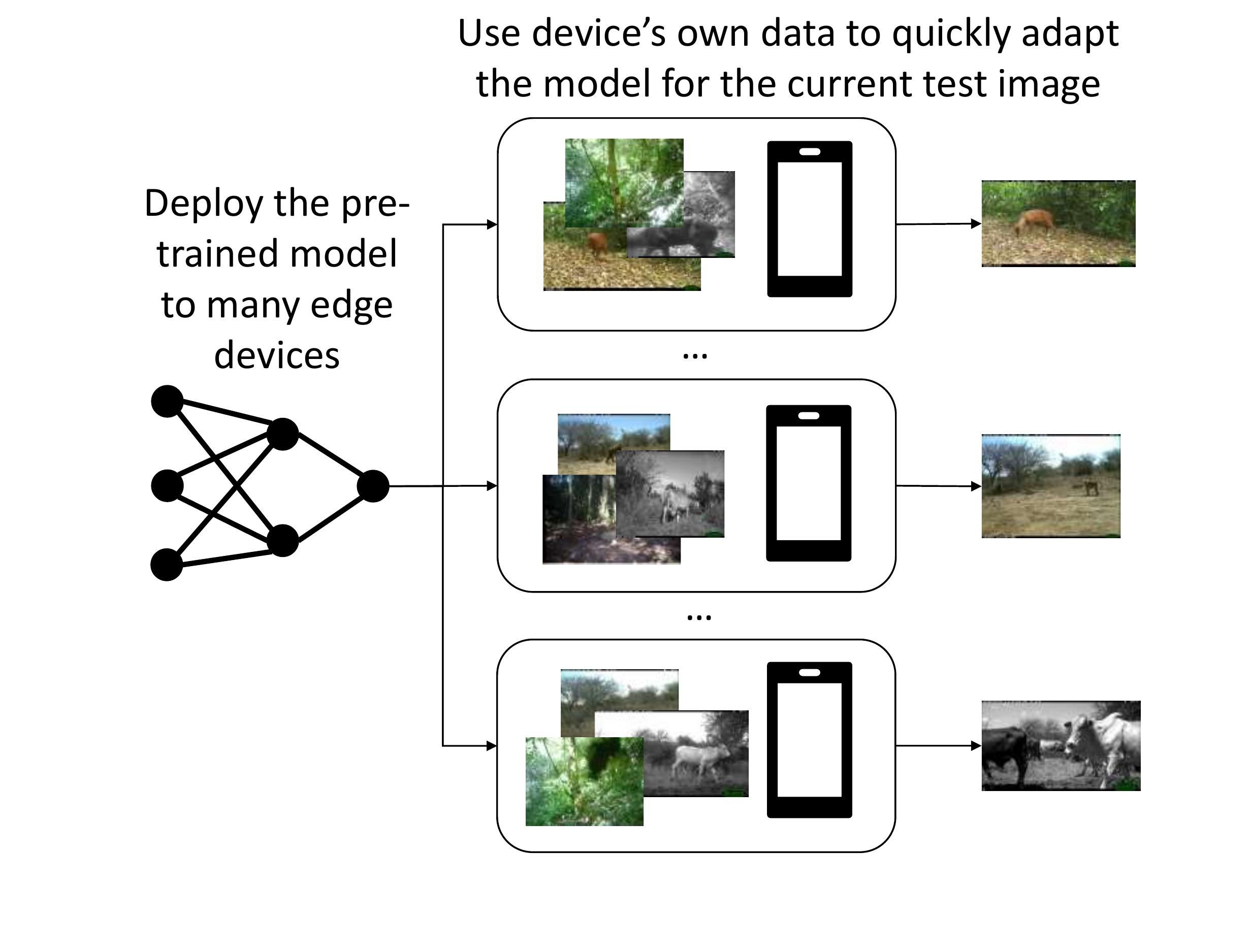}
  \vspace{-0.5cm}
  \caption{Illustration of the desired application scenario where a pre-trained model is deployed to many edge devices. Each device utilizes its own data coming from several domains to quickly adapt the model for the current test image.}
  \label{fig:devices}
\end{figure}

\keypoint{Our solution}
To solve the challenge posed earlier, we propose a feed-forward adaptation framework based on cross-attention between test instances and the adaptation set. The cross-attention module is meta-learned based on a set of training domains, inspired by \cite{Zhang2021AdaptiveShift}. This is a one-off cost paid up-front and performed on a server, after which the actual adaptation is fast. The pre-trained model is meant to be deployed to a large number of devices, each of which would benefit from being able to do fast feed-forward adaptation to its own unique data. Figure \ref{fig:devices} illustrates the desired application scenario. The deployed recognition model flexibly enables each inference operation to draw upon any part of the target adaptation set, exploiting each adaptation instance to a continuous degree. This can improve performance by eliminating adaptation instances that would be conventionally in-domain yet lead to negative transfer (e.g. same camera/opposite time of day), and include transfer from adaptation instances that would conventionally be out-of-domain but could benefit transfer (e.g. similar images/different camera).
Our experiments show that our cross-attention approach provides useful adaptation in this highly practical setting across a variety of synthetic and real benchmarks. 

\section{Background and related work}

\keypoint{Test-time domain adaptation} TTDA  has emerged as a topical adaptation scenario that focuses on model adaptation without access to source data at adaptation time (i.e. source-free condition), and further adapts to each minibatch at test time, aligning with an online adaptation scenario.  A meta-learning framework for TTDA has been recently proposed under the name adaptive risk minimization (ARM) \cite{Zhang2021AdaptiveShift}. ARM provides a variety of options for how TTDA is done, including context network that embeds information from the whole minibatch and updates to batch normalization statistics. ARM learns to do TTDA by meta-learning across a large number of tasks. TENT \cite{Wang2021Tent:Minimization} is another TTDA method and is based on optimizing channel-wise affine transformation according to the current minibatch. There is also continual TTDA (CoTTA) \cite{wang2022continualTTT} that considers non-stationary and continually changing environments. A common limitation of TTDA is that it typically requires adapting to a whole minibatch at once rather than one example at a time (e.g. to obtain more reliable statistics for adaptation). In contrast, in our setup, we can utilize device's own data to adapt the model to incoming test examples one at a time.

\keypoint{Latent domains} In settings with latent domains, information about domains is not available i.e. there are no domain labels. Further, some domains may be more similar to each other so the boundaries between domains are often blurred. Various approaches have been proposed to deal with latent domains, e.g. sparse latent adapters (SLA) \cite{Deecke2022VisualDomains}, domain agnostic learning \cite{Peng2019DomainRepresentations} that disentangles domain-specific features from class information using an autoencoder and \cite{Mancini2021InferringAdaptation} that discovers multiple latent domains using a specialized architecture with multiple branches. However, these methods focus on standard domain adaptation or supervised learning that is done across many epochs of training. Related setting is the open compound domain adaptation (OCDA) \cite{liu2020compoundDA, zhao2022sourcesegmentation}, where the target domain includes a mixture of unlabeled domains. Evaluation images come from one of the target domains or new unseen domains. Compared to our setup, OCDA is typically not source-free and uses back-propagation. However, SF-OCDA method \cite{zhao2022sourcesegmentation} that was developed for the problem of segmentation is source-free.

\keypoint{Meta-learning} Meta-learning can take various forms \cite{Hospedales2021Meta-learningSurvey}, but one of the most common is episodic meta-learning where the learning is done across tasks. As part of episodic meta-learning, the goal is to learn meta-parameters that will help achieve strong performance when presented with a new task. The most popular application of episodic meta-learning is few-shot learning, where we try to e.g. learn to distinguish between different classes after seeing only a few examples of each class. Example meta-parameters include weight initialization \cite{Finn2017Model-agnosticNetworks, Antoniou2019HowMAML, Li2017Meta-SGD:Learning} and metric space as is the case in Prototypical \cite{Snell2017PrototypicalLearning} and Relation networks \cite{Sung2018LearningLearning}. Prototypical and Relation networks are both examples of feed-forward methods. Episodic meta-learning has also been used in domain generalization \cite{Li2019EpisodicGeneralization} and as part of the ARM framework \cite{Zhang2021AdaptiveShift}. Episodic (task-based) meta-learning models our setup well because each deployment of the pre-trained model to a new device represents a new task, with its unique data to be used for adaptation. We use the episodic meta-learning paradigm.

\keypoint{Transformers} Our method uses cross-attention inspired by the attention mechanism found in the transformer architecture \cite{Vaswani2017AttentionNeed}. After transformers became common in natural language processing, they have also led to strong results within computer vision, most prominently as part of the ViT model \cite{Dosovitskiy2021AnScale}. ViT model has served as foundation for more recent vision transformers, including CrossViT \cite{Chen2021CrossViT:Classification} that combines strong performance with efficiency.
Our cross-attention mechanism is broadly inspired by the CrossViT cross-attention module \cite{Chen2021CrossViT:Classification}. Our approach has also been inspired by the idea of non-parametric transformers \cite{Kossen2021Self-attentionLearning} that can reason about relationships between data points. Different to CrossViT, we use image-to-image attention, instead of patch-to-patch, and show how to exploit this for feed-forward source-free latent domain adaptation. Cross-attention has also found use in few-shot classification \cite{Hou2019CrossClassification, Doersch2020CrossTransformers:Transfer, Ye2020Few-shotFunctions}, but these approaches use it to obtain better representations of class prototypes for few-shot supervised learning rather than to reason about which unlabelled examples come from relevant domains to drive unsupervised learning.

\section{Methods}
\subsection{Set-up}
\label{sec:set-up}
\keypoint{Preliminaries} During deployment the high-level assumption made by many source-free domain adaptation frameworks is that we are provided with a predictive model $f_\psi$ and an unlabeled target adaptation dataset $\sx$ whose label-space is the same as that of the pre-trained model \cite{Liang2020DoAdaptation}. Given these, source-free DA approaches define an algorithm $\mathcal{A}$ that ultimately leads to classifying a test instance $x_q$ as $y_q\approx\hat{y}_q=\mathcal{A}(x_q,\sx,\psi)$. There are numerous existing algorithms  for this. For example, pseudo-label strategies \cite{Liang2020DoAdaptation, Li2020ModelData, Yang2021ExploitingAdaptation} proceed by estimating labels $\hat{\boldsymbol{y}}_s$ for the adaptation set $\sx$, treating these as ground-truth, back-propagating to update the model $\psi'$ such that it predicts $\hat{\boldsymbol{y}}_s$, and then classifying the test point as $f_{\psi'}(x_q)$. We address the \emph{feed-forward} setting where algorithm $\mathcal{A}$ should not use back-propagation. For example, BN-based approaches \cite{Schneider2020ImprovingAdaptation,Zhang2021AdaptiveShift} use the adaptation set $\sx$ to update the BN statistics in $\psi$ as $\psi'$ and then classify the test point as  $f_{\psi'}(x_q)$. 

While the conventional domain adaptation setting assumes that target domain test examples $x_q$ and target domain training examples $\sx$ are all drawn from a common distribution, the \emph{latent domain} assumption has no such requirement. For example, $\sx$ may be drawn from a mixture distribution and $x_q$ may be drawn from only one component of that mixture. In this case only a subset of elements in $\sx$ may be relevant to adapting the inference for $x_q$. 

\keypoint{Deployment phase} 
Rather than explicitly updating model parameters, we aim to define a flexible inference routine $f_\psi$ that processes both $x_q$ and $\sx$ to produce $\hat{y}_q$ in a feed-forward manner, i.e. $\hat{y}_q=\mathcal{A}(x_q,\sx,\psi)=f_\psi(x_q,\sx)$. In this regard our inference procedure follows a similar flow to variants of ARM  \cite{Zhang2021AdaptiveShift}, with the following key differences: 1) ARM is transductive: it processes a batch of instances at once without distinguishing test instances and target adaptation set, so $\sx$ is composed of elements $x_q$. 2) ARM makes the conventional domain-observed assumption that domains have been defined by an external oracle that ensures all $x_q$ and $\sx$ are drawn from the same distribution.  We do not make this assumption and assess robustness to irrelevant elements in $\sx$. 

\keypoint{Pre-training phase} To train a model that can be used as described above, we follow an episodic meta-learning paradigm \cite{Zhang2021AdaptiveShift,Hospedales2021Meta-learningSurvey}. This refers to training $f_\psi$ using a set of simulated domain adaptation tasks. At each iteration, we generate a task with a unique pair of query and support instances  $(\sx,(y_q,x_q))$, keeping label space the same across all tasks. We simulate training episodes where $\sx$ contains instances with varying relevance to $x_q$. The goal is for $f_\psi$ to learn how to select and exploit instances from $\sx$ in order to adapt inference for $x_q$ to better predict $y_q$. 

In particular, our task sampler defines each task as having support examples uniformly sampled across a random set of $N_D$ domains, with the query example being from one of these domains. 
More formally, each task can be defined as:
$$\mathcal{T}=\left\{\left\{x_{s,1}, x_{s,2}, \dots x_{s, N_{s}}\right\}, x_q, y_q\right\}$$

for $N_s$ unlabelled support examples $x_{s,.}$ and query example $x_q$ with label $y_q$.

\keypoint{Sampling of latent domains} During pre-training and deployment, disjoint sets of domains are used, from which domains are sampled to create individual tasks. In each task the domain of the query example belongs to the set of support domains available on the deployed device. The support set contains also irrelevant domains to the query (test) set, but there are no domain labels in the task.

\subsection{Objective}
Our goal is to train a model that can adapt to relevant examples from the support set and obtain superior performance on the query examples. We can formalize this using the following objective:

\begin{equation}
\begin{split}
\min_{\featparams, \clsparams, \xaparams}&\mathcal{E}(\featparams,\clsparams,\xaparams)=\mathbb{E}_{p_d | \{d_i\}}\mathbb{E}_{p_{\{x_q, y_q \vert d\}, \{x_s | \{d_i\}\}}}\\ 
&\left[\frac{1}{N_q}\sum_{k=1}^{N_q} \ell (\cls(f_{\featparams \circ \xaparams} (x_{q,k};\sx),y_{q,k})) \right],\label{eq:obj}
\end{split}
\end{equation}

where $\boldsymbol{\psi}=\{\featparams, \clsparams, \xaparams\}$ are the parameters of the feature extractor, classifier and cross-attention module respectively (described in detail next), $\sx$ are the support examples used for adaptation, while $\qx$ are the query examples for which we make predictions and come from domain $d$. The support examples come from a set of domains $\{d_i\}$ with $d \in \{d_i\}$. There are $N_q$ query examples and $\mathcal{E}$ represents the generalization error after adapting the model.

\subsection{Architecture}

The key to solving Eq.~\ref{eq:obj} is defining an architecture $f_{\boldsymbol{\psi}}$ that can identify and exploit relevant support instances within $\sx$. Our solution to this relies on cross-attention between query and support images.
We first embed the support and query examples using the same feature extractor, after which we pass the embeddings through the cross-attention module. The output of cross-attention module is added to the embeddings of the query examples as a residual connection, after which the classifier makes predictions. Compared to CrossViT \cite{Chen2021CrossViT:Classification}, we do cross-attention between support and query images from different domains, image-to-image rather than patch-to-patch and on extracted features right before the classifier layer.

\keypoint{Cross-attention module}
Given a set of support examples $\sx$ and query examples $\qx$, we use the feature extractor $\feat$ to extract features $\feat(\sx)$, $\feat(\qx)$. Cross-attention module $\xa(\feat(\sx); \feat(\qx))$ parameterized by $\xaparams$ then transforms query embeddings $\feat(\qx)$, using support embeddings $\feat(\sx)$ as keys. The output is added to the query example features via a residual connection, which is then used by the classifier $\cls$ to predict labels of the query examples $\hat{\boldsymbol{y}}_q=\cls(\feat(\qx) + \xa(\feat(\sx); \feat(\qx)))$.

The cross-attention module itself performs image-to-image cross-attention, rather than patch-to-patch. More specifically, after extracting the features we flatten all spatial dimensions and channels into one vector, which represents the whole image.
The overall representation should intuitively better capture the nature of the domain rather than a patch, inspiring the use of image-to-image attention for domain adaptation. Performance comparison with patch-to-patch attention is included in the appendix.

Our cross-attention module is parameterized by a set of learnable projection matrices $\boldsymbol{W}_q, \boldsymbol{W}_k, \boldsymbol{W}_v$ (all of size $\mathbb{R}^{C\times (C/R)}$) with additional projection matrix $\boldsymbol{W} \in \mathbb{R}^{(C/R) \times C}$ to transform the queried outputs (we refer to all of these parameters collectively as $\boldsymbol{\omega}$). The output of the feature extractor $\feat$ is flattened into one vector (any spatial information is flattened), giving $C$ channels, so $\feat(\qx) \in \mathbb{R}^{N_q \times C}, \feat(\sx) \in \mathbb{R}^{N_s \times C}$. We also specify ratio $R$ that allows us to use rectangular projection matrices with fewer parameters, which improves efficiency and also provides regularization.

Formally we express $\xa$ as:
$$\boldsymbol{q}=\feat(\qx) \boldsymbol{W}_q, \quad \boldsymbol{k}=\feat(\sx) \boldsymbol{W}_k, \quad \boldsymbol{v}=\feat(\sx) \boldsymbol{W}_v,$$
$$\boldsymbol{A}=\texttt{softmax}\left(\boldsymbol{q}\boldsymbol{k}^T/\sqrt{C/h}\right), \quad \xa(\feat(\sx))=\boldsymbol{A}\boldsymbol{v}.$$

Similarly as CrossViT \cite{Chen2021CrossViT:Classification} and self-attention more broadly, we use multiple heads $h$, so we refer to it as $\texttt{MCA}$. We also use layer normalization as is the common practice. The output of $\texttt{MCA}$ is added to the query example embeddings as a residual connection:
$$\boldsymbol{z}=\feat(\qx) + \mxa(\texttt{LN}(\feat(\sx)); \texttt{LN}(\feat(\qx)))),$$
which is then passed through the classifier $\cls$ to obtain predictions $\hat{y}=\cls(\boldsymbol{z}).$ Following CrossViT, we do not apply a feed-forward network after cross-attention. We directly add the output via residual connection and pass it to the classifier.

\subsection{Meta-learning}
We train the main model (composed of the feature extractor $\feat$ and classifier $\cls$) and the cross-attention module (parameterized by $\xaparams$) by meta-learning across many tasks. Each task has the structure described in Section~\ref{sec:set-up}. Meta-learning is computationally efficient in this case because the inner loop does not include back-propagation based optimization -- the adaptation to the support examples is done purely feed-forward. We show how we do 
feed-forward adaptation on a new task combined with inference using \ourModel{} in Algorithm \ref{alg:inference}. Algorithm~\ref{alg:metalearningalgo} shows how we do the episodic pre-training that allows us to do efficient adaptation on new tasks during deployment.

\begin{algorithm}[t]
\small
\linespread{1.2}\selectfont
\caption{Feed-forward adaptation to a new task combined with inference using \ourModel{} }
\label{alg:inference}
\vspace{.25em}
\begin{algorithmic}[1]
    \REQUIRE{Model parameters $\featparams, \clsparams, \xaparams$, support $\sx$ and query $\qx$ examples from new domains}
    \STATE $\hat{\boldsymbol{y}}_q\leftarrow\cls(\feat(\qx) + \mxa(\texttt{LN}(\feat(\sx)); \texttt{LN}(\feat(\qx)), \xaparams)))$
\end{algorithmic}
\end{algorithm}

\begin{algorithm}[t]
\small
\linespread{1.2}\selectfont
\caption{Episodic pre-training for source-free latent domain adaptation with \ourModel{} }
\label{alg:metalearningalgo}
\vspace{.25em}
\begin{algorithmic}[1]
    \REQUIRE{\# training steps $T$, \# latent domains in a task $N_D$, \# support examples $N_s$, \# query examples $N_q$, learning rate $\eta$}
    \STATE \textbf{Initialize:} $\featparams, \clsparams, \xaparams$
    \FOR{$t=1,\ldots,T$}
        \STATE Sample $N_D$ support domains $\{\mathbb{D}_s\}_1^{N_D}$ from training domains
        \STATE Sample query domain $\mathbb{D}_q$ from the support domains $\{\mathbb{D}_s\}_1^{N_D}$
        \STATE Sample $N_s$ unlabelled support images $\sx$ uniformly from the selected support domains $\{\mathbb{D}_s\}_1^{N_D}$ 
        \STATE Sample $N_q$ labelled query images $\qx, \boldsymbol{y}_q$ from domain $\mathbb{D}_q$
        \STATE Predict query labels $\hat{\boldsymbol{y}}_q\leftarrow\cls(\feat(\qx) + \mxa(\texttt{LN}(\feat(\sx)); \texttt{LN}(\feat(\qx)))))$
        \STATE $(\featparams,\clsparams,\xaparams)\leftarrow(\featparams,\clsparams,\xaparams)-\eta\nabla_{(\featparams,\clsparams,\xaparams)}\sum_{k=1}^{N_q}\ell(\hat{y}_{q,k}, y_{q,k})$
    \ENDFOR
\end{algorithmic}
\end{algorithm}

\section{Experiments}
\subsection{Benchmarks}
We evaluate our approach on a variety of synthetic and real-world benchmarks, namely FEMNIST \cite{Caldas2018LEAF:Settings}, CIFAR-C \cite{Hendrycks2019BenchmarkingPerturbations}, TinyImageNet-C \cite{Hendrycks2019BenchmarkingPerturbations} and iWildCam \cite{Beery2020TheDataset}. These benchmarks have a large number of domains, e.g. around 100 for CIFAR-C and TinyImageNet-C, and around 300 for FEMNIST and iWildCam. Using a large number of domains for pre-training is reasonable as for many practical problems it is possible to collect such pre-training data. We describe each benchmark next.

\textbf{FEMNIST} dataset includes images of handwritten letters and digits, and is derived from the EMNIST dataset \cite{Cohen2017EMNIST:Letters} by treating each writer as a domain. \textbf{CIFAR-C} extends CIFAR-10 \cite{Krizhevsky2009LearningImages} by a applying a variety of corruptions such as different brightness, snow or various types of blurring. There are different levels of severity with which the corruptions are applied, giving rise to multiple domains for the different levels. \textbf{TinyImageNet-C} is an extension of TinyImageNet analogous to CIFAR-C. \textbf{iWildCam} is a large-scale real-world dataset that includes images of different animal species taken by cameras in different locations. There is a lot of variability in the style of images in different cameras, for example different illumination, camera angle or vegetation. The dataset has also substantial class imbalance, so macro F1 score is used for evaluation.

For FEMNIST, CIFAR-C and TinyImageNet-C we follow the splits into meta-training, meta-validation and meta-testing sets as selected in \cite{Zhang2021AdaptiveShift}. For iWildCam we follow the splits of domains selected in \cite{koh2021wilds}. Additionally for iWildCam we filter out domains with fewer than 40 examples.

\subsection{Baselines}
\keypoint{ERM} Empirical risk minimization or ERM is a domain generalization baseline that simply trains on all training domains and performs no domain adaptation. It is known to work surprisingly well and is often difficult to beat when properly tuned \cite{Gulrajani2020InGeneralization}. In our case it is trained following the episodic pipeline for fair comparison i.e. it is directly trained using the query examples during meta-training.

\keypoint{BN} A simple and often useful method for source-free domain adaptation is to update the batch normalization statistics using the unlabelled target domain data \cite{Schneider2020ImprovingAdaptation}. It has achieved strong results in conventional source-free domain adaptation (SFDA) \cite{Ishii2021Source-freeStatistics}. However, in the latent DA setting it is unclear if statistics calculated across a support set of varying relevance will be helpful for achieving better performance. During evaluation, the statistics are updated using all support examples, and directly used for the query examples.

\keypoint{CML} Contextual Meta-Learning is the main instantiation of ARM  \cite{Zhang2021AdaptiveShift} as a way to extract information from the whole minibatch in test-time adaptation and use it to obtain better performance on test images. We apply the CML on the whole support set with images from different domains and then use it as additional information for making predictions on test images. CML is a feed-forward domain adaptation method, but it has not been designed for the latent domain adaptation problem.

\keypoint{Back-prop-based} Fine-tuning with standard domain adaptive losses such as entropy minimisation \cite{grandvalet2004sslEntropyMin} (FT-EM) and infomax \cite{shi2012unsupDA} (FT-IM) for 10 steps with 0.1x of standard learning rate. Fine-tuning with these objectives is widely used in prior SFDA methods, and these baselines roughly correspond to applying methods such as TENT \cite{Wang2021Tent:Minimization} and SHOT \cite{Liang2020DoAdaptation} to our problem setting respectively. 
These results are presented for context rather than fair comparison, because they use back-prop (which we do not) and they are not designed for latent domain adaptation (unlike us). 

\keypoint{Latent and continual DA approaches} We have repurposed several methods designed to handle latent or continually changing domains into our setup to test more advanced baselines. The methods we evaluate are SF-OCDA \cite{zhao2022sourcesegmentation}, CoTTA \cite{wang2022continualTTT} and SLA \cite{Deecke2022VisualDomains}. SF-OCDA and CoTTA use back-propagation for the adaptation, while SLA uses adapters to handle new domains. Further details are in the appendix.

\subsection{Implementation details}
\keypoint{Our solution -- \ourModel{}}
Our cross-attention module first flattens all spatial information and channels into one vector for each image, so it works image-to-image. In line with existing literature \cite{Vaswani2017AttentionNeed, Chen2021CrossViT:Classification}, we use 8 heads and layer normalization on the flattened features of support and query images. The use of layer normalization means that our approach does not rely on a minibatch of query examples i.e. it natively supports streaming mode and does not need mutiple query examples to obtain strong results, unlike existing test-time domain adaptation approaches \cite{Zhang2021AdaptiveShift, Wang2021Tent:Minimization}.

Support images are projected into keys and values, while query images act as queries for cross-attention after transformation by a projection matrix. After calculating the attention map and applying it to the values, we multiply the output by a further projection matrix. We use only one cross-attention layer and our projection matrices have rectangular shape of $C \times C/2$ where $C$ is the dimensionality of the flattened features. No dropout is used.

\keypoint{Data augmentation} 
We use weak data augmentation during meta-training. The exact augmentations are cropping, horizontal flipping, small rotations (up to 30 degrees) and are different from the corruptions tested in some of the benchmarks. These are applied with probability 0.5 independently.

\keypoint{Task sampling}
Our tasks have 5 support domains, with 20 examples in each, overall 100 support examples (realistic for a deployed device). Query examples come from one randomly selected support set domain (out of 5 options) and there are 20 of them. Note that the method fully supports the streaming mode, so it works independently for each query example without having to calculate  statistics across the minibatch. The exact number of tasks for meta-validation and meta-testing is respectively (420, 420) for FEMNIST, (850, 11000) for CIFAR-C, (1700, 11000)  for TinyImageNet-C, and (745,  2125)  for iWildCam.

\keypoint{Training}
We use hyperparameters from \cite{Zhang2021AdaptiveShift} for FEMNIST, CIFAR-C and TinyImageNet-C, and we also train the cross-attention parameters with the same optimizer. For FEMNIST and CIFAR-C a small CNN model is used, while for TinyImageNet-C a pre-trained ResNet-50 \cite{He2015DeepRecognition} is fine-tuned. For iWildCam we also fine-tune a pre-trained ResNet50 model, and we follow the hyperparameters selected in \cite{koh2021wilds}, but with images resized to $112\times 112$ and training for 50 epochs. 
All our experiments are repeated across three random seeds.

\newcommand{\tc}[1]{\multicolumn{2}{c}{#1}}
\begin{table*}[t]
\begin{center}
\begin{adjustbox}{max width=0.90\textwidth}
\begin{tabular}{lcccccccc}
    \toprule
& \tc{\textbf{FEMNIST}} & \tc{\textbf{CIFAR-C}} & \tc{\textbf{ TinyImageNet-C}} & \tc{\textbf{ iWildCam}} \\
  \cmidrule(lr){2-3}                    \cmidrule(lr){4-5}                    \cmidrule(lr){6-7} \cmidrule(lr){8-9} 
\textbf{Approach} & \textbf{W10\%} & \textbf{Avg} & \textbf{W10\%} & \textbf{Avg} & \textbf{W10\%} & \textbf{Avg} & \textbf{W10\%} & \textbf{Avg} \\
\midrule
ERM & 52.7 $\pm$ 1.4 & 77.2 $\pm$ 0.9 & 44.3 $\pm$ 0.5 & 68.6 $\pm$ 0.3 & 4.8 $\pm$ 0.2  & 26.4 $\pm$ 0.4 & 0.0 $\pm$ 0.0 & 38.7 $\pm$ 0.8 \\
CML \cite{Zhang2021AdaptiveShift} & 50.4 $\pm$ 1.3 & 76.0 $\pm$ 0.9 & 44.8 $\pm$ 0.5 & 69.5 $\pm$ 0.5 & 4.8 $\pm$ 0.5 & 25.7 $\pm$ 0.6 & 0.0 $\pm$ 0.0 & 38.7 $\pm$ 1.1 \\
BN \cite{Ishii2021Source-freeStatistics,Zhang2021AdaptiveShift}& 52.2 $\pm$ 1.5 & 78.0 $\pm$ 0.7 & 45.4 $\pm$ 0.7 & 69.3 $\pm$ 0.4 & 5.9 $\pm$ 0.2 & 27.7 $\pm$ 0.3 & 1.9 $\pm$ 1.1 & 42.5 $\pm$ 0.8 \\
Our \ourModel{} & \textbf{53.3 $\pm$ 0.6} & \textbf{78.3 $\pm$ 0.0} & \textbf{49.4 $\pm$ 0.6}  & \textbf{72.0 $\pm$ 0.3} & \textbf{6.5 $\pm$ 0.2} & \textbf{28.6 $\pm$ 0.3} & \textbf{3.6 $\pm$ 1.5} & \textbf{43.5 $\pm$ 1.5} \\
\midrule
FT-EM (TENT) \cite{grandvalet2004sslEntropyMin} & 51.7 $\pm$ 1.4 & 77.6 $\pm$ 0.8 & 44.9 $\pm$ 0.6 & 69.2 $\pm$ 0.4 & 3.9 $\pm$ 0.4 & 25.7 $\pm$ 0.3 & 0.0 $\pm$ 0.0 & 38.6 $\pm$ 0.8 \\
FT-IM (SHOT) \cite{shi2012unsupDA,Liang2020DoAdaptation} & 52.5 $\pm$ 1.2 & 77.5 $\pm$ 0.8 & 45.6 $\pm$ 0.5 & 69.5 $\pm$ 0.3 & 4.8 $\pm$ 0.4 & 24.6 $\pm$ 1.0 & 0.0 $\pm$ 0.0 & 38.7 $\pm$ 0.8\\
\midrule
SF-OCDA \cite{zhao2022sourcesegmentation} & 51.5 $\pm$ 1.4 &  77.5 $\pm$ 0.7 & 46.7 $\pm$ 1.2 & 70.1 $\pm$ 0.5 & 5.5 $\pm$ 0.2  & 26.7 $\pm$ 0.2 & 0.0 $\pm$ 0.0 & 38.4 $\pm$ 0.6 \\
CoTTA \cite{wang2022continualTTT} & 51.4 $\pm$ 0.4 & 76.8 $\pm$ 0.2 & 46.2 $\pm$ 0.3  & 69.8 $\pm$ 0.2 & 4.9 $\pm$ 0.5 & 26.0 $\pm$ 0.7 & 0.0 $\pm$ 0.0 & 38.6 $\pm$ 0.5  \\
SLA \cite{Deecke2022VisualDomains} & 46.0 $\pm$ 1.4 & 74.1 $\pm$ 0.8 & 40.8 $\pm$ 1.1 & 64.0 $\pm$ 0.7 & 2.5 $\pm$ 0.1 & 16.9 $\pm$ 0.3 & 0.0 $\pm$ 0.0 & 29.9 $\pm$ 1.4 \\
\bottomrule
  \end{tabular}
  \end{adjustbox}
\end{center}
  \caption{Main benchmark results: average and worst-case (worst 10\% tasks) test performance, with standard error of the mean across 3 random seeds. Accuracy is reported for all except iWildCam, where F1 score is used (\%). The best results are highlighted in bold. Our \ourModel{} approach achieves the best performance across all of the benchmarks.}
  \label{tab:main}
\end{table*}

\keypoint{Evaluation metrics} We follow \cite{Zhang2021AdaptiveShift} in reporting average and worst performance over all testing tasks. While \cite{Zhang2021AdaptiveShift} reports the worst single task, we modify this metric to report the average performance of the worst decile of tasks. The reason is that for some benchmarks, among many test tasks with varying domain transfer difficulty there can easily be at least one task with zero accuracy.

\subsection{Results}
We report our results in Table \ref{tab:main} for all benchmarks: FEMNIST, CIFAR-C, TinyImageNet-C and large-scale real-world iWildCam benchmark. We include both average performance as well as reliability via the worst case performance \cite{Zhang2021AdaptiveShift}, with our bottom decile modification. 
From the results we can see our cross-attention approach results in consistent improvements over the strong ERM baseline across all benchmarks, as well as the other baselines. The encouraging result on iWildCam highlights our method works well also in practical real-world scenarios.

Overall we see CML and BN strategies that naively combine information from all support examples have limited success when the support set has both domain relevant and domain irrelevant examples. In fact, CML achieves lower performance than ERM in some of the benchmarks, despite having a mechanism for domain adaptation. The results show the need to adaptively select the right examples from the support set when they come from domains of mixed relevance. The results confirm our cross-attention based framework can successfully select useful information from the set of examples with both relevant and irrelevant examples and ultimately achieve superior performance. Back-propagation-based alternatives usually perform worse, despite being slower, due to lack of support for latent domains. Methods for latent or continual domain adaptation have also not obtained strong results, likely because the setups they were designed for are too far from our setup -- e.g. we do not focus on segmentation and we use a source-free setup.

\subsection{Further analysis}
As part of analysis we study several questions: 1) How does the performance of unsupervised cross-attention compare with a supervised version? 2) How fast is our approach in comparison with the other approaches? 3) What do the attention weights look like and what do they imply? 4) How does the performance vary with variable number of domains in the support sets? 5) How does the size of the support set influence the performance?

\keypoint{Domain-supervised adaptation} Recall that our main \ourModel{} algorithm and experiments earlier are domain unsupervised. This may induce a cost due to distraction by domain-irrelevant adaptation data (e.g. as observed by CML under-performing ERM previously) or a potential benefit due to enabling transfer.
We therefore compare our unsupervised method with a domain-supervised alternative, with manually defined attention weights based on domain labels. Table~\ref{tab:sup-ca-main} shows the results are dataset-dependent. The fact that in at least some cases domain-unsupervised adaptation outperforms the supervised case shows that the benefit can sometimes outweigh the cost, and that it is possible for a suitable model to outperform manual domain annotations.

\begin{table*}[h!]
\begin{center}
\begin{adjustbox}{max width=0.56\textwidth}
\begin{tabular}{lcccc}
    \toprule
Cross-attention & FEMNIST & CIFAR-C & TinyImageNet-C & iWildCam \\
\midrule
Domain-unsupervised & 78.3 $\pm$ 0.0 & 72.0 $\pm$ 0.3 & 28.6 $\pm$ 0.3 & 43.5 $\pm$ 1.5 \\
Domain-supervised & 79.4 $\pm$ 0.4 & 69.8 $\pm$ 0.4 & 28.6 $\pm$ 0.2 & 52.0 $\pm$ 1.2 \\
\bottomrule
  \end{tabular}
\end{adjustbox}
\end{center}
  \caption{Comparison of domain-unsupervised and domain-supervised CXDA on our benchmarks. Average test accuracy for all benchmarks apart from iWildCam where F1 score is reported (\%). Domain supervision is helpful in multiple cases, but can be outperformed.}
  \label{tab:sup-ca-main}
\end{table*}

\begin{figure*}[h!]
 \begin{center}
 \includegraphics[width=\linewidth]{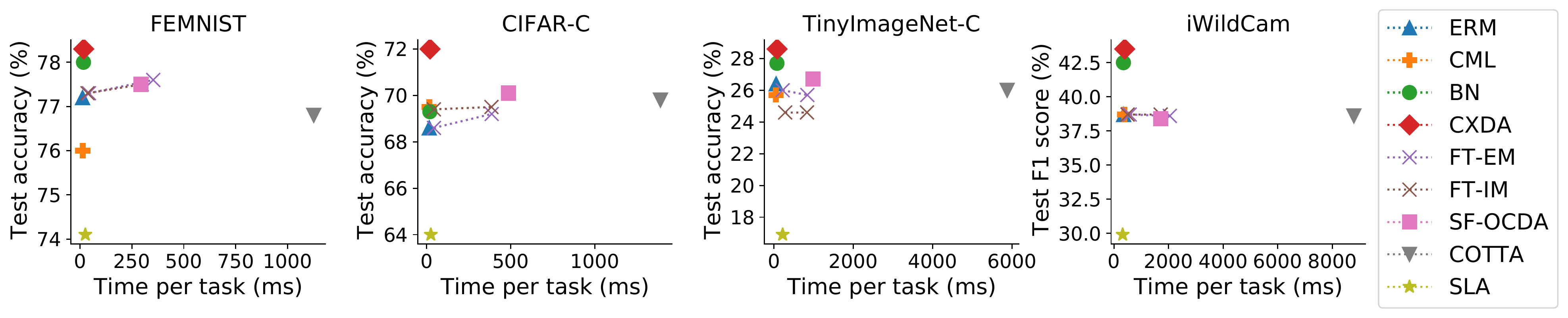}
  \caption{Analysis of test accuracy (\%) vs time per task (ms) for the various approaches evaluated. \ourModel{} achieves the best performance, has similar speed to other feed-forward baselines and is faster than fine-tuning approaches that use back-propagation (1 and 10 adaptation steps are shown for FT-EM and FT-IM). The difference is especially large when the fine-tuning approaches use 10 fine-tuning steps, but even if only 1 step is used there is a visible speed difference. Time per task includes adapting to the task and making a prediction.}
  \label{fig:ft-variable-steps-all}
 \end{center}
\end{figure*}

\keypoint{Speed evaluation}
We compare the accuracy vs time per task (adaptation and inference) in Figure \ref{fig:ft-variable-steps-all}. The analysis shows our \ourModel{} approach achieves the best performances, is capable of real-time adaptation with similar speed as the other feed-forward baselines, and is significantly faster than the back-propagation based approaches. The precise times are in Table~\ref{tab:bothTime} in the appendix, together with the time needed for pre-training. While our method requires more time for pre-training than the baselines, this cost is amortized and not of interest due to the focus on being fast when solving new adaptation tasks during deployment.

\begin{figure}
\begin{center}
  \includegraphics[width=\linewidth]{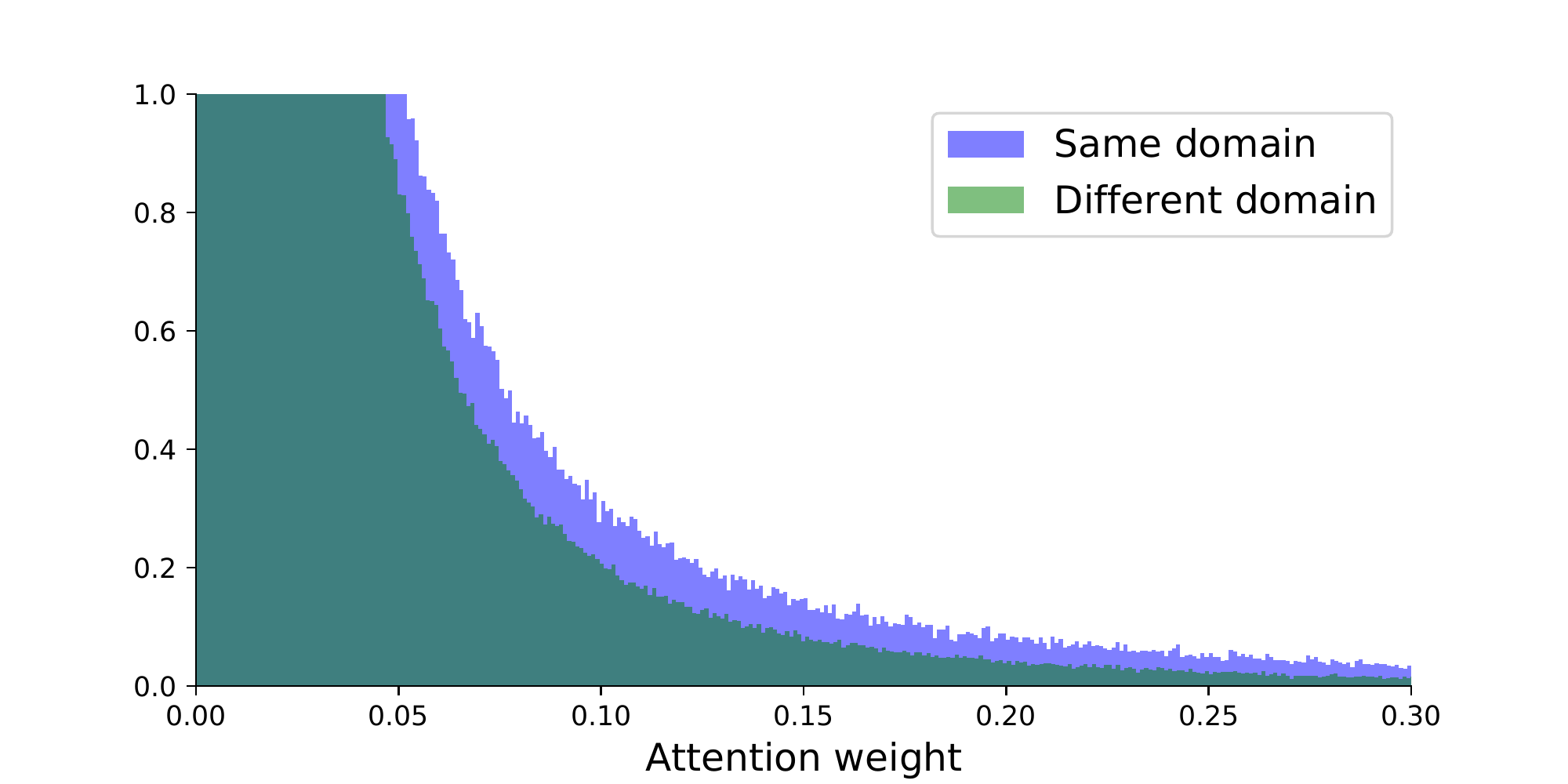}
  \caption{Density histograms of attention weights for pairs of same and different domain examples in the test tasks of iWildCam.}
  \label{fig:iwcattns}
    \vspace{-0.65cm}
\end{center}
\end{figure}

\keypoint{Analysis of attention weights} We have analysed the attention weights to understand the learned behaviour of the cross-attention mechanism. We have selected the large-scale iWildCam benchmark and used one of the trained cross-attention models. Figure \ref{fig:iwcattns} shows the density histogram of attention weights for same and different domain support examples, relative to the query examples in each task. From the plot we observe: 1) There is a significant amount of weight spent on attending to examples in domains different from the current query. This suggests that the model is exploiting knowledge transfer beyond the boundaries of the standard (camera-wise) domain annotation in the benchmark, as illustrated in Figure~\ref{fig:ldatask}. 2) Nevertheless, overall the weight distribution tends to attend more strongly to the in-domain instances than out-of-domain instances. This shows that our cross-attention module has successfully learned how to match query instances with corresponding domain instances in the support set, despite never experiencing domain-supervision.

\keypoint{Variable number of domains} Tables \ref{tab:var_domains_w10} and \ref{tab:var_domains} in the appendix show that the best performance is obtained when there are fewer domains, confirming our intuition. However, \ourModel{} can handle well also cases when there is a large number of domains and consistently outperforms other approaches irrespective of the number of domains.

\keypoint{Variable support set size} Tables \ref{tab:var_sizes_w10} and \ref{tab:var_sizes} in the appendix analyse the impact of variable number of examples in the support set. The results confirm \ourModel{} is scalable and outperforms other approaches even if the support set size changes to more or fewer examples.

\subsection{Discussion}
\keypoint{Broader impact} \ourModel{} is designed to help obtain better performance in settings which are underrepresented in the training data. 
It performs adaptation when back-propagation is not supported, and thus benefits users whose resources may be limited to lower-power, embedded or mobile devices.

\keypoint{Scalability and applicability} Our cross-attention approach is fast in the scenario when there are a moderate number of examples for adaptation (in our experiments we use 100), and not too many inferences to be made per adaptation. In this regime it is much faster than the mainstream line of back-propagation-based adaptation solutions \cite{kouw2021daReview,csurka2022visual,Liang2020DoAdaptation,Wang2021Tent:Minimization}. Whereas, if there are very many inferences to be made, then the overhead cost of back-propagation would eventually be amortized. As for attention more broadly, its computational cost depends on the number of query and support examples,
and the approach would become increasingly expensive in computation if the number of support examples became very large. For a very large support set, one could simply use a random or semi-random subset of images for adaptation. However, the most effective  use case is one where the adaptation set is smaller and/or changes rapidly compared to the frequency of inference. Instantiations of our approach using efficient approximations to attention can be developed as part of future work.

\section{Conclusion}
We have introduced a new highly practical setting suitable for resource-constrained devices, where we efficiently adapt a pre-trained model using examples that come from a mixture of domains and are without domain or class labels.
To answer this new challenging adaptation problem, we have developed a novel solution based on cross-attention that is able to automatically select relevant examples and use them for fast feed-forward adaptation that happens in real time.

\clearpage

{\small
\bibliographystyle{ieee_fullname}
\bibliography{references,morerefs}
}

\appendix

\clearpage

\section{Additional details}
We have followed the experimental choices from \cite{Zhang2021AdaptiveShift} for FEMNIST, CIFAR-10-C, TinyImageNet-C and \cite{koh2021wilds} for iWildCam, unless we specify otherwise. This includes the splits of data into training, validation and test sets as well as the splits of domains into meta-training, meta-validation and meta-test sets of domains. For iWildCam we use the OOD splits from \cite{koh2021wilds}.

\subsection{Models}

Feature extractor and classifier:
\begin{itemize}
    \item FEMNIST: CNN with three convolutional layers, hidden dimension of 128, batch normalization, ReLU activation, kernel of 5, padding of 2. Classifier consists of two fully-connected layers with 200 hidden units and ReLU activation in between. The input shape of images is $28\times 28$.
    \item CIFAR-C: Same architecture as for FEMNIST, but with three input channels instead of 1 (colour images). The input shape of images is $32\times 32$.
    \item TinyImageNet-C: ImageNet pre-trained ResNet50 \cite{He2015DeepRecognition}. The classifier consists of one fully connected layer. The input shape of images is $64\times 64$.
    \item iWildCam: ImageNet pre-trained ResNet50 \cite{He2015DeepRecognition}. The classifier consists of one fully connected layer. The input shape of images is $112\times 112$.
\end{itemize}

Adaptation-specific components:
\begin{itemize}
    \item CML: Context network is used to transform the support examples -- three convolutional layers, 64 hidden units, kernel size of 5, padding of 2, with batch normalization and ReLU activation. The output of the network has the same shape as input. To create the context we average the context network outputs across the support examples and use the same context for all query examples in the task.
    \item CXDA: All key details are explained in the main text, and we provide more detailed explanations here. The size of the fully-connected layers depends on the flattened shape of the features -- only one cross-attention layer is used. After multiplying attention weights with projected values ($\boldsymbol{A}\boldsymbol{v}$), we transform the output further using projection matrix $\boldsymbol{W}$, similarly as \cite{Chen2021CrossViT:Classification}. However, we do not use a further MLP model that would contain multiple layers and non-linearity, so we also follow \cite{Chen2021CrossViT:Classification} in this aspect. The output of the cross-attention module has the same shape as input. As part of CXDA, batch normalization statistics of the feature extractor are updated too using the support set.
\end{itemize}

\subsection{Training}
Dataset-specific training details:
\begin{itemize}
    \item FEMNIST: SGD with learning rate of $10^{-4}$, momentum of 0.9 and weight decay of $10^{-4}$, trained for 200 epochs, with validation set evaluated every 10 epochs, and early stopping based on accuracy.
    \item CIFAR-C: SGD with learning rate of $10^{-2}$, momentum of 0.9 and weight decay of $10^{-4}$, trained for 100 epochs, with validation set evaluated every 10 epochs, and early stopping based on accuracy.
    \item TinyImageNet-C: SGD with learning rate of $10^{-2}$, momentum of 0.9 and weight decay of $10^{-4}$, trained for 50 epochs, with validation set evaluated every 5 epochs, and early stopping based on accuracy.
    \item iWildCam: Adam with learning rate of $3\times 10^{-5}$, no weight decay, trained for 50 epochs, with validation set evaluated every 5 epochs, and early stopping based on macro F1 score.
\end{itemize}

In all cases we use cross-entropy loss, and the cross-attention parameters are optimized in the same way as the main model. In each iteration we use a task that has 5 domains with 20 support examples for each sampled domain, and there are 20 query examples from one selected domain from the set of current domains.

Details about fine-tuning (FT) of the pre-trained ERM model: we perform fine-tuning by taking 10x smaller learning rate compared to the ones used during training and then performing 10 steps on the task's support data. We consider two losses: 1) entropy minimization (EM) and 2) information maximization (IM) of the support set example predictions during the fine-tuning. We reset the model to the pre-trained state for each test task.

\subsection{Latent and continual domain adaptation}

\keypoint{SF-OCDA} When using SF-OCDA, pre-training is the same as for ERM, but the adaptation on evaluation tasks is specialized. As part of SF-OCDA \cite{zhao2022sourcesegmentation} repurposed into our setup, we perform 10 update steps with 10x smaller learning rate, similar to our other back-propagation based baselines. Following \cite{zhao2022sourcesegmentation}, we use cross-entropy loss between the pseudo-labels and the predicted labels, which are predicted for clean support data or augmented support data with 70\% and 30\% probability respectively. The augmentations are more advanced and include color jitter, random affine transformation, Gaussian blur, random horizontal flip and Gaussian noise.

\keypoint{CoTTA} Similar to other back-propagation based approaches, only the adaptation to evaluation tasks is unique and uses 10 update steps with 10x smaller learning rate than used during pre-training. During adaptation we directly follow \cite{wang2022continualTTT} and use the support examples for adaptation. The key idea of the method is to use weight-averaged and augmentation-averaged predictions in order to reduce error accumulation. Additionally a small random part of neurons is restored to the pre-trained weights in each iteration, which helps prevent catastrophic forgetting.

\keypoint{SLA} SLA \cite{Deecke2022VisualDomains} modifies the architecture of the main model to include gates and corrections (adapters) to handle latent domains. The adapters are trained alongside the main model in the pre-training stage. Since already two adapters are shown to perform well in \cite{Deecke2022VisualDomains}, we also use two of them. Support examples of evaluation tasks are not used for adaptation because the model with trained adapters is directly used to make predictions on the query examples, with the adaptation done by using the adapters.

\section{Qualitative analysis}
\begin{figure*}[h!]
  \begin{center}
  \includegraphics[width=0.1\linewidth]{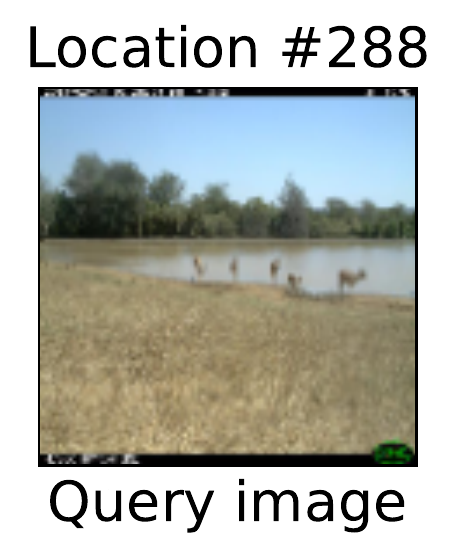}
  \includegraphics[width=\linewidth]{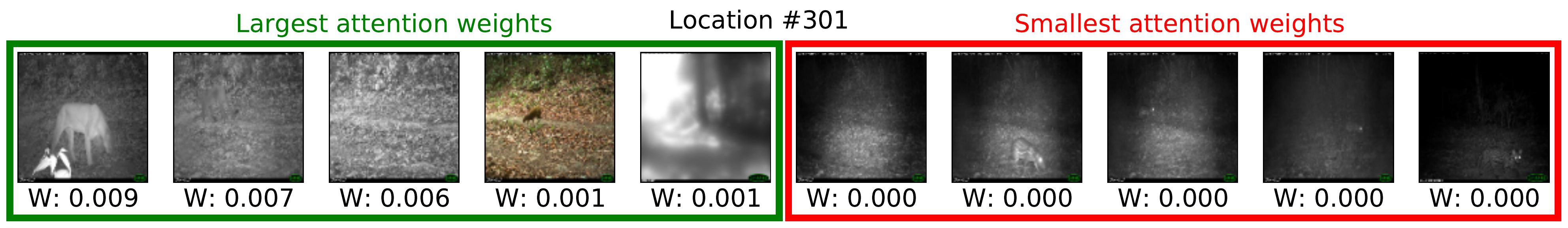}
  \includegraphics[width=\linewidth]{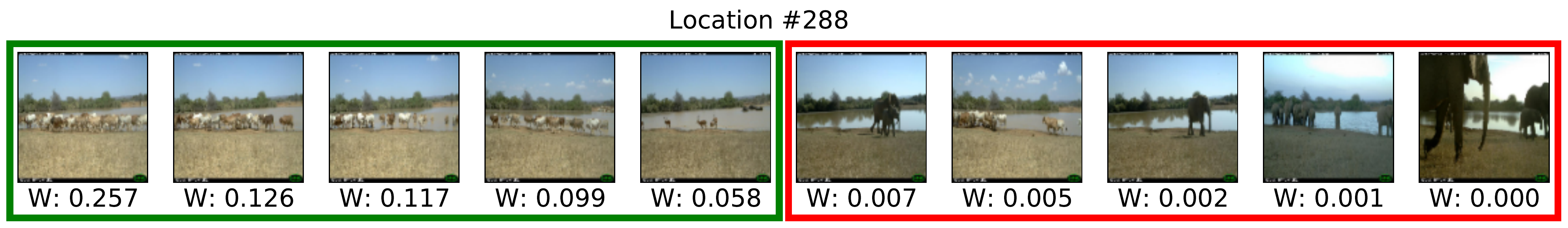}
  \includegraphics[width=\linewidth]{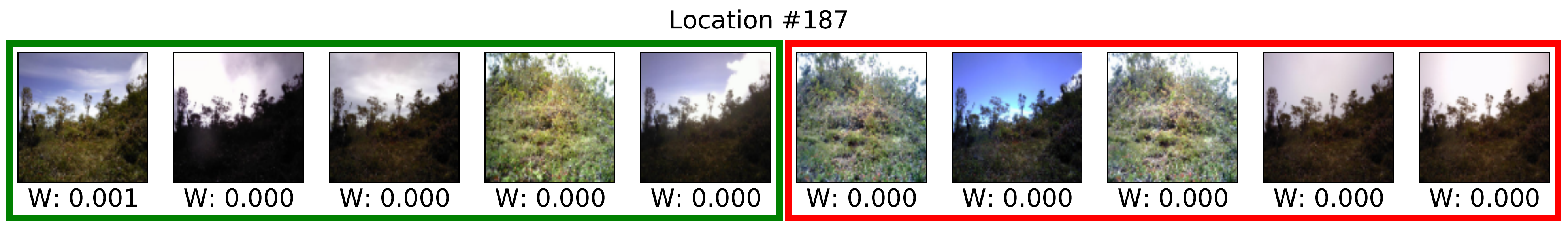}
  \includegraphics[width=\linewidth]{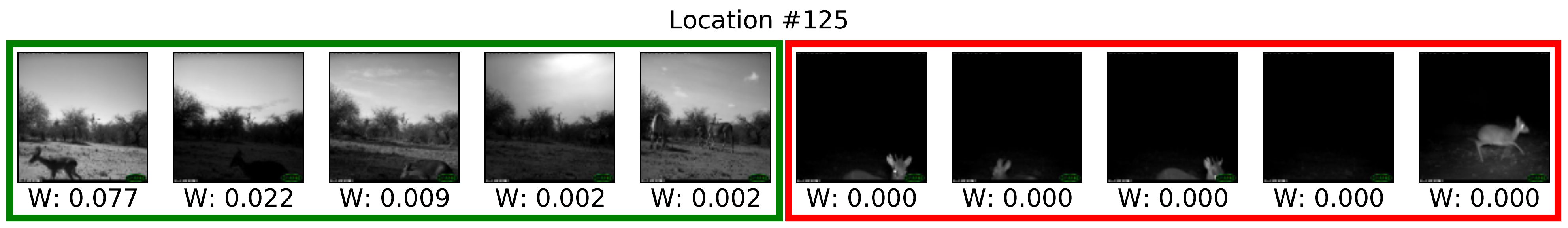}
  \includegraphics[width=\linewidth]{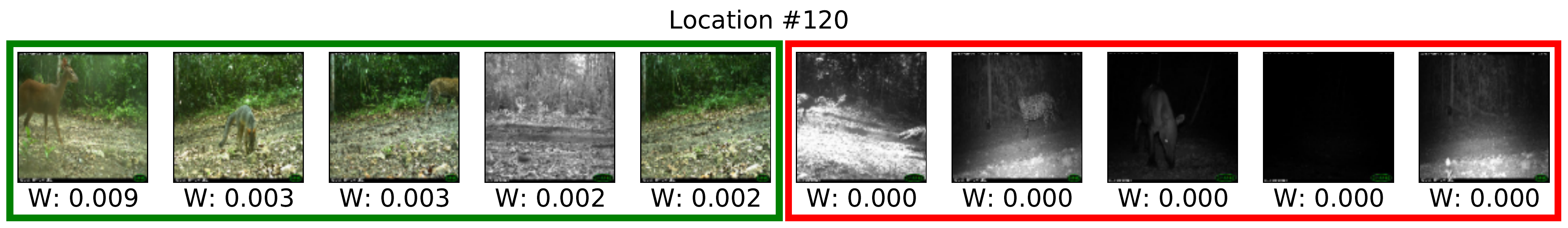}
  \caption{Analysis of attention weights for an example task in iWildCam, with a query image coming from location (camera trap) \#288. We show the five support examples in each domain that have the largest and smallest attention weights. Similar images from the same location (\#288) are given the largest weights, but also relevant images from other locations (e.g. \#125) are given larger weights. The examples with the smallest attention weights visually do not seem relevant.}
  \label{fig:qualitative-analysis}
  \end{center}
\end{figure*}

We provide qualitative analysis of our approach in Figure \ref{fig:qualitative-analysis}. The analysis shows that similar images from the same location are given the largest weights, but also relevant images from other locations are given larger weights.

\section{Runtime analysis}
We provide a more detailed runtime analysis in Table \ref{tab:bothTime}.
Table~\ref{tab:bothTime} (left) shows that the time per task (combination of adaptation and inference) of \ourModel{} is very similar to feed-forward baselines and significantly faster than the time of fine-tuning (back-propagation) baselines. Table \ref{tab:bothTime} (right) shows that the episodic pre-training takes longer for the smaller datasets, but the difference is small for large datasets and models. However, the pre-training time is not of particular interest to us because of the focus on fast adaptation to new tasks during deployment. Overall back-propagation-based approaches have significantly larger inference times than the feed-forward ones, showing the need for specialized feed-forward adaptation methods. We additionally note the total runtime of fine-tuning methods is closer to the feed-forward approaches because their training is done in a standard way. All experiments within the same benchmark used the same GPU and number of CPUs.

\begin{table*}[h!]
\begin{center}
\begin{adjustbox}{max width=\textwidth}
\begin{tabular}{lcccccccc}
    \toprule
    & \multicolumn{4}{c}{Time for adaptation and inference on a task (ms/task)} & \multicolumn{4}{c}{Total runtime for pre-training and adaptation (minutes)} \\
Approach & FEMNIST & CIFAR-C & TinyImageNet-C & iWildCam & FEMNIST & CIFAR-C & TinyImageNet-C & iWildCam \\
\midrule
ERM & 12.0 $\pm$ 0.1 & 15.7 $\pm$ 0.3 & 52.8 $\pm$ 0.2 & 352.0 $\pm$ 2.8 & \phantom{0}38.3 $\pm$ \phantom{0}0.3 & 26.6 $\pm$ 0.2 & \phantom{0}94.1 $\pm$ 0.2 & 440.2 $\pm$ 5.4\\
CML & 13.1 $\pm$ 0.2 & 15.8 $\pm$ 0.1 & 48.9 $\pm$ 0.1 & 385.2 $\pm$ 8.0 & \phantom{0}55.2 $\pm$ \phantom{0}1.0 & 35.2 $\pm$ 0.2 & 131.8 $\pm$ 0.2 & 461.9 $\pm$ 7.2 \\
BN & 16.5 $\pm$ 0.7 & 19.2 $\pm$ 0.2 & 74.1 $\pm$ 0.1 & 345.2 $\pm$ 1.9 & \phantom{0}44.5 $\pm$ \phantom{0}0.3 & 31.6 $\pm$ 0.3 & 112.3 $\pm$ 0.3 & 432.2 $\pm$ 0.8 \\
\ourModel{} & 17.8 $\pm$ 1.5 & 20.5 $\pm$ 0.2 & 77.9 $\pm$ 2.8 & 392.5 $\pm$ 1.3 & 110.0 $\pm$ 12.0 & 63.2 $\pm$ 0.6 & 167.9 $\pm$ 1.9 & 491.6 $\pm$ 1.0 \\
\midrule
FT-EM & 352.7 $\pm$ 42.4 & 387.1 $\pm$ 10.5 & 840.2 $\pm$ 15.4 & 2044.7 $\pm$ 238.0 & 97.5 $\pm$ 6.4 & 156.6 $\pm$ 3.6 & 483.4 $\pm$ 4.9 & 907.1 $\pm$ 28.0 \\
FT-IM & 296.9 $\pm$ \phantom{0}3.1 & 385.9 $\pm$ 10.9 & 830.7 $\pm$ 16.6 & 1709.8 $\pm$ \phantom{0}16.5 & 78.2 $\pm$ 0.4 & 150.2 $\pm$ 3.9 & 470.8 $\pm$ 7.3 & 688.0 $\pm$ 14.8 \\
\midrule
SF-OCDA  & \phantom{0}294.0 $\pm$ \phantom{00}7.5 & \phantom{0}488.0 $\pm$ \phantom{00}8.8 & \phantom{0}980.8 $\pm$ \phantom{00}8.8 & 1708.6 $\pm$ \phantom{00}1.9 & \phantom{0}77.7 $\pm$ \phantom{0}1.5 & 229.5 $\pm$ 34.3 & \phantom{0}630.8 $\pm$ 23.8 & \phantom{0}599.8 $\pm$ \phantom{0}3.1 \\
CoTTA & 1126.2 $\pm$ 177.4 & 1390.9 $\pm$ \phantom{0}20.1 & 5876.5 $\pm$ 196.8 & 8790.9 $\pm$ 335.7 & 209.3 $\pm$ 29.1 & 492.2 $\pm$ \phantom{0}7.0 & 3009.4 $\pm$ 93.4 & 1823.5 $\pm$ 85.8 \\
SLA & \phantom{00}25.5 $\pm$ \phantom{00}0.8 & \phantom{00}25.1 $\pm$ \phantom{00}0.0 & \phantom{0}217.6 $\pm$ \phantom{00}0.5 & \phantom{0}317.3 $\pm$ \phantom{00}5.2 & \phantom{0}85.5 $\pm$ \phantom{0}2.3 & \phantom{0}49.7 $\pm$ \phantom{0}0.1 & \phantom{0}367.9 $\pm$ \phantom{0}3.9 & \phantom{0}489.4 $\pm$ \phantom{0}5.8 \\
\bottomrule
  \end{tabular}
\end{adjustbox}
\end{center}
  \caption{Comparative computational cost of different adaptation methods for adaptation and pre-training.}
  \label{tab:bothTime}
\end{table*}

\section{Additional analyses}
\label{app:analyses}
The additional analyses use pre-trained models that were trained with tasks that have 100 support examples coming from 5 domains. We then deploy them to tasks that have 1) variable number of support domains or 2) variable support set sizes. Tables \ref{tab:var_domains_w10} and \ref{tab:var_domains} show that the best performance is obtained when there are fewer domains and that \ourModel{} can handle well also cases when there is a large number of domains.
Tables \ref{tab:var_sizes_w10} and \ref{tab:var_sizes} confirm \ourModel{} is scalable and outperforms other approaches even if the support set size changes to more or fewer examples. Since the latent and continual domain adaptation methods have not shown to be promising in the main setup, we do not evaluate them as part of the additional analyses.

\begin{table*}[h!]
\begin{center}
\begin{adjustbox}{max width=\textwidth}\begin{tabular}{lcccccc}
    \toprule
Approach & 1 domain & 2 domains & 5 domains & 10 domains & 20 domains \\
\midrule
ERM & 58.3 $\pm$ 0.4 & 54.2 $\pm$ 0.3 & 44.3 $\pm$ 0.5 & 37.9 $\pm$ 0.5 & 29.4 $\pm$ 0.4 \\
CML & 57.7 $\pm$ 0.7 & 54.3 $\pm$ 0.6 & 44.8 $\pm$ 0.5 & 39.2 $\pm$ 0.4 & 30.6 $\pm$ 0.4 \\
BN & 59.7 $\pm$ 0.5 & 55.4 $\pm$ 0.5 & 45.4 $\pm$ 0.7 & 38.9 $\pm$ 0.7 & 30.2 $\pm$ 0.6 \\
\ourModel{} & 62.7 $\pm$ 0.4 & 58.7 $\pm$ 0.3 & 49.4 $\pm$ 0.6 & 43.0 $\pm$ 0.5 & 33.2 $\pm$ 0.3 \\
\midrule
FT-EM & 49.4 $\pm$ 0.6 & 49.0 $\pm$ 0.5 & 44.9 $\pm$ 0.6 & 39.1 $\pm$ 0.5 & 30.5 $\pm$ 0.4 \\
FT-IM & 53.1 $\pm$ 0.5 & 50.7 $\pm$ 0.5 & 45.6 $\pm$ 0.5 & 39.6 $\pm$ 0.5 & 30.7 $\pm$ 0.4 \\
\bottomrule
  \end{tabular}
  \end{adjustbox}
\end{center}
  \caption{Analysis of the impact of variable number of domains in the support set -- worst 10\% tasks test accuracy on CIFAR-C (\%).}
  \label{tab:var_domains_w10}
\end{table*}

\begin{table*}[h!]
\begin{center}
\begin{adjustbox}{max width=\textwidth}\begin{tabular}{lcccccc}
    \toprule
Approach & 1 domain & 2 domains & 5 domains & 10 domains & 20 domains \\
\midrule
ERM & 68.7 $\pm$ 0.3 & 68.7 $\pm$ 0.3 & 68.6 $\pm$ 0.3 & 68.5 $\pm$ 0.2 & 68.6 $\pm$ 0.2 \\
CML & 68.8 $\pm$ 0.6 & 69.2 $\pm$ 0.5 & 69.5 $\pm$ 0.5 & 69.4 $\pm$ 0.4 & 69.5 $\pm$ 0.4 \\
BN & 69.7 $\pm$ 0.4 & 69.4 $\pm$ 0.4 & 69.3 $\pm$ 0.4 & 69.1 $\pm$ 0.4 & 69.2 $\pm$ 0.4 \\
\ourModel{} & 72.2 $\pm$ 0.2 & 72.1 $\pm$ 0.2 & 72.0 $\pm$ 0.3 & 71.9 $\pm$ 0.3 & 71.9 $\pm$ 0.3 \\
\midrule
FT-EM & 69.0 $\pm$ 0.4 & 69.1 $\pm$ 0.3 & 69.2 $\pm$ 0.4 & 69.2 $\pm$ 0.3 & 69.3 $\pm$ 0.3 \\
FT-IM & 69.8 $\pm$ 0.3 & 69.6 $\pm$ 0.3 & 69.5 $\pm$ 0.3 & 69.4 $\pm$ 0.3 & 69.5 $\pm$ 0.3 \\
\bottomrule
  \end{tabular}
\end{adjustbox}
\end{center}
\caption{Analysis of the impact of variable number of domains in the support set -- average test task accuracy on CIFAR-C (\%).}
  \label{tab:var_domains}
\end{table*}

\begin{table*}[h!]
\begin{center}
\begin{adjustbox}{max width=\textwidth}\begin{tabular}{lccccccc}
    \toprule
Approach & 10 examples & 20 examples & 50 examples & 100 examples & 200 examples & 500 examples \\
\midrule
ERM & 0.1 $\pm$ 0.0 & 21.7 $\pm$ 0.1 & 38.2 $\pm$ 0.5 & 44.3 $\pm$ 0.5 & 48.2 $\pm$ 0.6 & 50.2 $\pm$ 0.7 \\
CML & 0.0 $\pm$ 0.0 & 21.7 $\pm$ 0.2 & 38.9 $\pm$ 0.5 & 44.8 $\pm$ 0.5 & 48.6 $\pm$ 0.5 & 50.6 $\pm$ 0.4 \\
BN & 1.2 $\pm$ 1.0 & 22.8 $\pm$ 0.9 & 39.5 $\pm$ 0.7 & 45.4 $\pm$ 0.7 & 49.4 $\pm$ 0.7 & 51.3 $\pm$ 0.7 \\
\ourModel{} & 1.8 $\pm$ 0.9 & 26.4 $\pm$ 0.7 & 43.0 $\pm$ 0.5 & 49.4 $\pm$ 0.6 & 53.8 $\pm$ 0.6 & 56.2 $\pm$ 0.8 \\
\midrule
FT-EM & 0.8 $\pm$ 0.7 & 22.4 $\pm$ 0.6 & 39.0 $\pm$ 0.5 & 44.9 $\pm$ 0.6 & 48.9 $\pm$ 0.5 & 50.8 $\pm$ 0.5 \\
FT-IM & 1.3 $\pm$ 0.9 & 22.9 $\pm$ 0.8 & 39.7 $\pm$ 0.5 & 45.6 $\pm$ 0.5 & 49.6 $\pm$ 0.5 & 51.7 $\pm$ 0.4 \\
\bottomrule
  \end{tabular}
  \end{adjustbox}
\end{center}
  \caption{Analysis of the impact of variable number of examples in the support set -- worst 10\% tasks test accuracy on CIFAR-C (\%).}
  \label{tab:var_sizes_w10}
\end{table*}

\begin{table*}[h!]
\begin{center}
\begin{adjustbox}{max width=\textwidth}\begin{tabular}{lccccccc}
    \toprule
Approach & 10 examples & 20 examples & 50 examples & 100 examples & 200 examples & 500 examples \\
\midrule
ERM & 68.7 $\pm$ 0.3 & 68.7 $\pm$ 0.3 & 68.6 $\pm$ 0.3 & 68.6 $\pm$ 0.3 & 68.6 $\pm$ 0.3 & 68.6 $\pm$ 0.3 \\
CML & 67.7 $\pm$ 0.3 & 68.7 $\pm$ 0.4 & 69.3 $\pm$ 0.4 & 69.5 $\pm$ 0.5 & 69.5 $\pm$ 0.5 & 69.6 $\pm$ 0.5 \\
BN & 69.3 $\pm$ 0.4 & 69.3 $\pm$ 0.4 & 69.2 $\pm$ 0.4 & 69.3 $\pm$ 0.4 & 69.2 $\pm$ 0.4 & 69.3 $\pm$ 0.4 \\
\ourModel{} & 69.6 $\pm$ 0.3 & 70.9 $\pm$ 0.3 & 71.7 $\pm$ 0.3 & 72.0 $\pm$ 0.3 & 72.1 $\pm$ 0.2 & 72.2 $\pm$ 0.3 \\
\midrule
FT-EM & 69.1 $\pm$ 0.3 & 69.2 $\pm$ 0.3 & 69.1 $\pm$ 0.3 & 69.2 $\pm$ 0.4 & 69.2 $\pm$ 0.3 & 69.2 $\pm$ 0.3 \\
FT-IM & 69.4 $\pm$ 0.3 & 69.5 $\pm$ 0.3 & 69.4 $\pm$ 0.3 & 69.5 $\pm$ 0.3 & 69.4 $\pm$ 0.3 & 69.5 $\pm$ 0.3 \\
\bottomrule
  \end{tabular}
  \end{adjustbox}
\end{center}
  \caption{Analysis of the impact of variable number of examples in the support set -- average test task accuracy on CIFAR-C (\%).}
  \label{tab:var_sizes}
\end{table*}

\section{Patch-to-patch attention}
We provide additional comparison of image-to-image and patch-to-patch (P2P) attention in Table \ref{tab:p2pattn}. Compared to standard image-to-image CXDA, CXDA P2P performs slightly worse, likely because using whole images is simpler and provides useful regularization. In terms of time, the detailed comparison depends on the setup, but overall both need relatively similar time in practice.

\begin{table*}[t]
\begin{center}
\begin{adjustbox}{max width=0.85\textwidth}
\begin{tabular}{lcccccccc}
    \toprule
& \tc{\textbf{FEMNIST}} & \tc{\textbf{CIFAR-C}} & \tc{\textbf{ TinyImageNet-C}} & \tc{\textbf{ iWildCam}} \\
  \cmidrule(lr){2-3}                    \cmidrule(lr){4-5}                    \cmidrule(lr){6-7} \cmidrule(lr){8-9} 
\textbf{Approach} & \textbf{W10\%} & \textbf{Avg} & \textbf{W10\%} & \textbf{Avg} & \textbf{W10\%} & \textbf{Avg} & \textbf{W10\%} & \textbf{Avg} \\
\midrule
ERM & 52.7 $\pm$ 1.4 & 77.2 $\pm$ 0.9 & 44.3 $\pm$ 0.5 & 68.6 $\pm$ 0.3 & 4.8 $\pm$ 0.2  & 26.4 $\pm$ 0.4 & 0.0 $\pm$ 0.0 & 38.7 $\pm$ 0.8 \\
\ourModel{} & 53.3 $\pm$ 0.6 & 78.3 $\pm$ 0.0 & 49.4 $\pm$ 0.6  & 72.0 $\pm$ 0.3 & 6.5 $\pm$ 0.2 & 28.6 $\pm$ 0.3 & 3.6 $\pm$ 1.5 & 43.5 $\pm$ 1.5 \\
\ourModel{} P2P & 52.0 $\pm$ 0.7 & 77.3 $\pm$ 0.3 &  45.6 $\pm$ 0.3 & 69.8 $\pm$ 0.2 & 6.8 $\pm$ 0.2 & 28.9 $\pm$ 0.1 & 4.1 $\pm$ 0.9 & 42.9 $\pm$ 1.3 \\
\bottomrule
  \end{tabular}
  \end{adjustbox}
\end{center}
  \caption{Comparison of image-to-image and patch-to-patch (P2P) attention: average and worst-case (worst 10\% tasks) test performance, with standard error of the mean across 3 random seeds. Accuracy is reported for all except iWildCam, where F1 score is used (\%). Our image-to-image \ourModel{} performs better than the patch-to-patch alternative in general.}
  \label{tab:p2pattn}
\end{table*}

\end{document}